\documentclass[letterpaper, 10 pt, conference]{ieeeconf}

\IEEEoverridecommandlockouts 

\usepackage{amsmath} 
\usepackage{amssymb}  
\newtheorem{remark}{Remark}
\newcommand{\norm}[1]{\left\lVert#1\right\rVert}
\usepackage{graphicx}

\usepackage{multirow}

\title{\LARGE \bf
Safety Considerations in Deep Control Policies with Safety Barrier Certificates Under Uncertainty
}
\author{Tom Hirshberg$^{1*}$, Sai Vemprala$^{2}$ and Ashish Kapoor$^{2}$
\thanks{*Work done while interning at Microsoft Corporation, Redmond}
\thanks{$^{1}$CS Department, Technion - Israel Institute of Technology, Haifa, Israel {\tt\small tomhirshberg@campus.technion.ac.il}}%
\thanks{$^{2}$Microsoft Corporation, Redmond, WA {\tt\small \{sai.vemprala, akapoor\}@microsoft.com}}%
}

\begin{document}

\maketitle
\thispagestyle{empty}
\pagestyle{empty}

\begin{abstract}
Recent advances in Deep Machine Learning have shown promise in solving complex perception and control loops via methods such as reinforcement and imitation learning. However, guaranteeing safety for such learned deep policies has been a challenge due to issues such as partial observability and difficulties in characterizing the behavior of the neural networks. While a lot of emphasis in safe learning has been placed during training, it is non-trivial to guarantee safety at deployment or test time. This paper extends how under mild assumptions, Safety Barrier Certificates can be used to guarantee safety with deep control policies despite uncertainty arising due to perception and other latent variables. Specifically for scenarios where the dynamics are smooth and uncertainty has a finite support, the proposed framework wraps around an existing deep control policy and generates safe actions by dynamically evaluating and modifying the policy from the embedded network. Our framework utilizes control barrier functions to create spaces of control actions that are safe under uncertainty, and when the original actions are found to be in violation of the safety constraint, uses quadratic programming to minimally modify the original actions to ensure they lie in the safe set. Representations of the environment are built through Euclidean signed distance fields that are then used to infer the safety of actions and to guarantee forward invariance. We implement this method in simulation in a drone-racing environment and show that our method results in safer actions compared to a baseline that only relies on imitation learning to generate control actions.
\end{abstract}

\section{INTRODUCTION}
Trained deep control policies via reinforcement learning (RL) and imitation learning (IL) allow generating control outputs directly from sensor inputs. However, in contrast to simulations and games, applying such techniques to real-world safety-critical applications remains an incredibly challenging task. The strong reliance on deep neural networks makes them vulnerable to overconfident or unpredictable results when presented with data distributions unseen during training. Additionally, in real-life scenarios there might be environmental factors (e.g. friction, wind, viscosity, etc.) and uncertainties due to machine perception, that may not have been explicitly modeled in the formulation. The importance of predictions and actions being robust to such exogenous variations is paramount in the safety-critical aspects of real-world robotics.

Much of the previous work on safety in deep control policies has focused on modifying the training phase. These include reward engineering \cite{long2018towards}, constrained optimization to incorporate safety constraints \cite{bouton2019safe} and worst-case optimization \cite{tang2019worst}. Providing safety guarantees that would hold at the deployment phase (test time) is challenging, since it is difficult to characterize or enumerate the complete state space of the agent. For example, it is impossible to characterize a-priori all images a robot would see. Furthermore, the mathematical structure of the deep policies further makes it difficult to provide an analysis of the deep policies.

In this work, we explore a runtime alternative that aims to keep the system safe by providing minimal deviations of control signals stemming from an embedded deep control policy. The framework attempts to continuously preserve the safety via a {\em barrier} function, while the agent continues to make progress towards the task it was trained for. In particular, the work extends Safety Barrier Certificates (SBC) \cite{Wang_2017} to handle safety considerations in deep control policies, and focuses solely on a test-time implementation, thus not affecting the training phase. We specifically focus on the problem of autonomous drone-racing, where a quadrotor needs to negotiate several gates without collisions while moving as fast as possible on a racing track.
There are several technical challenges that we encounter and address. First, deep control policies are often used when there is no explicit model of systems dynamics available. In the absence of such a dynamics model, it is non-trivial to use SBCs. Second, the safety constraints for applications such as drone-racing can be complex. For example, in our application case, the quadrotors are not allowed to collide with the gates or other objects in the environment. Representations such as Euclidean signed distance fields (ESDF) are popular and useful to formally define the safety conditions. However, it is unclear how SBCs can be applied here. Finally, the safety framework also needs to account for any uncertainty and non-determinism that might arise due to environmental factors.

The core insight in this work is that for many real-world applications system dynamics are well approximated as an ordinary differential equation that is uniformly continuous, bounded and Lipschitz continuous. This allows us to make a locally linear approximation while ensuring that the approximation error is small. Similarly, we incorporate safety constraints defined over ESDFs via smooth approximations and finally discuss safety under uncertainty and non-determinism. We implement our framework in simulation and show that our method results in guaranteeable safety and improved avoidance compared to the original deep control policies, even under perception uncertainty. In summary, the main contributions of this paper include:\\
\begin{itemize}
\item Enhancing any trained deep control policy using SBC under uncertainty.
\item Simplifying safety constraints for barrier function-based avoidance in complex environments.
\item Improving the safety of deep control policies.
\end{itemize}

\section{RELATED WORK}
Recent research has focused on learning control policies directly from raw data using deep neural networks (DNN) by either imitation or reinforcement learning \cite{pan2018agile}, \cite{lillicrap2015continuous}. Much of the work in safe deep control focuses on training time and aims to induce risk-aversion via reward function or through constrained optimization \cite{long2018towards, garcia2012safe, achiam2017constrained}. However, none of these approaches guarantee safety during the test or deployment phase. Formal verification and certification has also been proposed to address the application of deep neural networks in safety-critical applications. For example, \cite{katz2017reluplex, liu2019algorithms} focus on verification procedure of DNNs through analysis of activation functions and layers. Similarly, \cite{dutta2018learning} perform verification for a feedback control network using a receding horizon formulation that attempts to enforce properties such as reachability, safety and stability. \cite{fisac2019bridging} discuss control-theoretic modifications to reinforcement learning for safety analysis. The notion of probabilistic safety under uncertainty has also been explored previously via formal methods \cite{Sadigh-RSS-16}. Much of this work results in computationally intensive procedures that cannot be easily used in real-time systems.

Safety Barrier Certificates (SBC) with permissive control barrier functions (CBF), have been previously used to guarantee runtime safety in both deterministic and non-deterministic settings {\cite{ames2016control, Ames_2019, Wang_2017}}. CBF were also used for a safe exploration during learning of RL models \cite{cheng2019end, ohnishi2019barrier, li2019temporal}. The key idea is to first define a barrier function by considering a set of unsafe states and the system dynamics, and then use it to minimally modify a given controller so that the resulting solution is safe. The framework can be extended to handle uncertainty in the environment to probabilistically guarantee safety \cite{Wenhao2019}. 
Recent work by \cite{bajcsy2019efficient} also proposed a real-time safety framework on top of learning-based planners, based on Hamilton-Jacobi reachability. This paper builds upon this line of works where the key idea centers on wrapping a deep control policy within the SBC framework. However, unlike most applications of SBC, in our case there is no explicit system dynamics model available. The use of CBF to ensure safety was also introduced in \cite{chen2017obstacle, yang2019sampling}, which enforced CBF constraints and obtained control actions through a mixed integer program or a quadratic program. These works assumed that the obstacles were convex, whereas our framework can handle non-convex obstacles in a fast and easy way, forming a QP to be solved for CBF obstacle avoidance. In \cite{shi2019neural}, the authors present a deep controller that performs autonomous drone landing while providing stability guarantees under unsteady dynamics. In comparison, we explicitly also focus on observation uncertainty along with system uncertainty: as in most of the cases where deep controllers are applied there is uncertainty that stems from the sensing and perception part of the system.

We demonstrate the framework on the task of autonomous drone-racing, where within a realistic simulator \cite{Shah_2017}, a quadrotor uses an RGB camera to perceive and negotiate multiple gates as fast as possible on a racing track. The approaches to solve drone-racing consider inferring simple representation of the environment, and then using either classical control and planning methods \cite{kaufmann2019beauty, Li_2019} or building deep controllers \cite{pmlr-v87-kaufmann18a, bonatti2019learning}. In this paper we explore deep control policies and assume that at least one gate is always in the field of view of the quadrotor. 
\begin{figure*}[t]
    \begin{tabular}{p{0.7\textwidth}p{0.2\textwidth}}
    \hspace{0.2mm} 
    \includegraphics[scale=0.8]{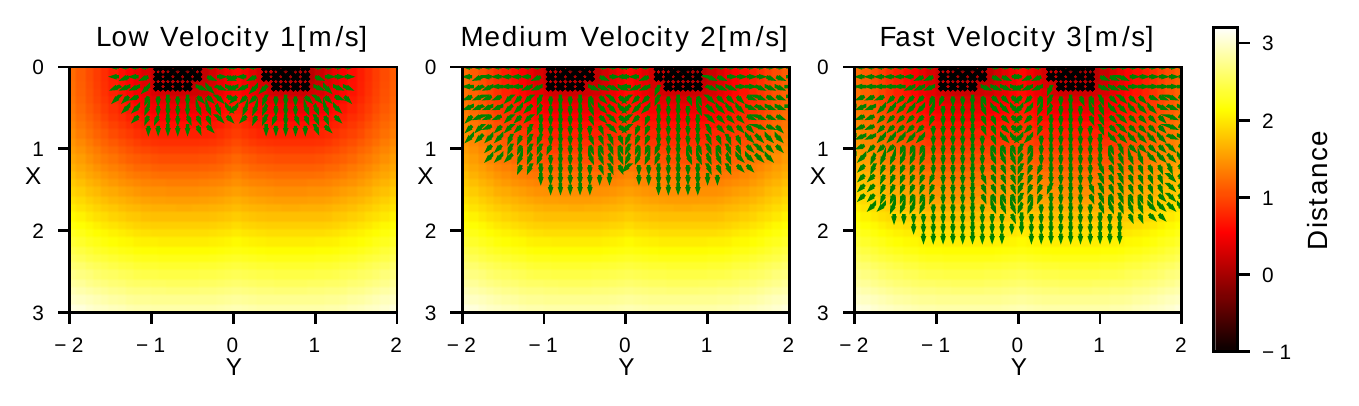}
    \label{fig:sdf_map11}&
    \multirow{2}{\hsize}[24mm]{
    \includegraphics[scale=0.3]{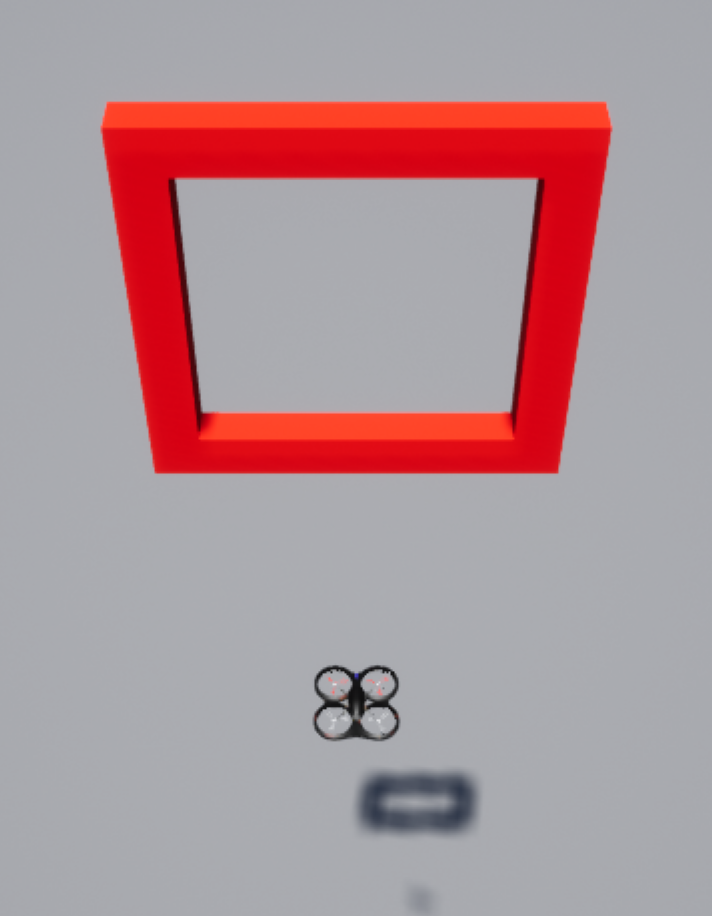}\label{fig:sdf_map12} }
    \\[-4mm]
    \includegraphics[scale=0.8]{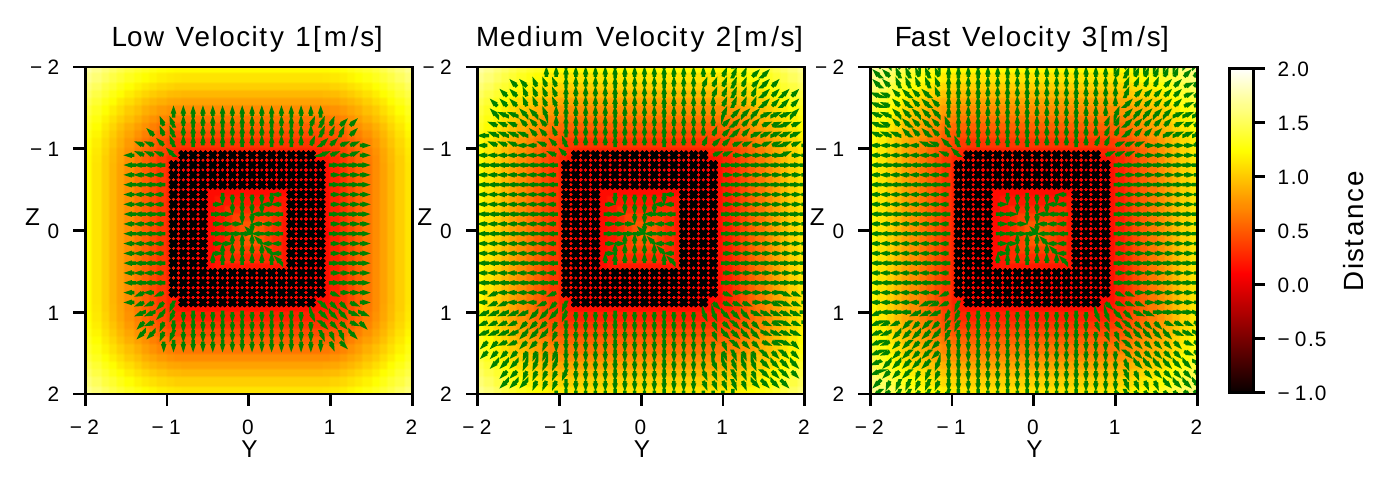}\label{fig:sdf_map13}&
    \end{tabular}
    \caption{The distance map defined by (\ref{eq:dist_func}), by $\mathbf{XY}$ and $\mathbf{YZ}$ planes (upper row and the lower row). The reader is guided to zoom-in the figures for details: The green arrows direct the safest action as computed by the SBC and providing the safety constraints, and the black crosses indicate positions that are unsafe in every direction. The actions sampled for every position and for every angle are $|\mathbf{u}|=velocity$. The right image shows the original gate.}
    \label{fig:sdf_map}
\end{figure*}
\section{PROPOSED FRAMEWORK}
Assume we have been given a policy, that produces a control signal ${\bf \tilde{u}}$. The goal of the proposed framework is to provide a projection of ${\bf\tilde{u}}$, such that the system is safe with respect to safety constraints. In the context of this discussion, we make use of an important assumption that the uncertainties pertaining to system dynamics as well as the perception observations can be modeled as distributions with finite support. The finite support assumption is obtained from knowledge of the application domain. As the system of interest is a quadrotor, due to physical actuation limits, the uncertainties arising from translational and rotational dynamics are bounded. At the same time, the perception based localization system that is responsible for generating control actions only attempts to localize the drone gate within the viewing frustum of the camera, hence bounding the observation uncertainty accordingly.

In order to use the SBC framework, we need to first characterize (1) system dynamics, (2) safety constraints and (3) handle uncertainty. We describe these in detail below:
\subsection{System Dynamics Model}
Due to the absence of an explicit system dynamics model in most deep control scenarios, we need to make certain assumptions. First, we consider a simplified dynamics model for a robot evolving as a continuous-time system:
\begin{align}
\dot{\mathbf{x}} &= f(\mathbf{x},\mathbf{u})+\mathbf{w}, \quad \mathbf{w}\sim U (-\Delta \mathbf{w},\Delta \mathbf{w}) \notag\\
\hat{\mathbf{x}} &= \mathbf{x}+\mathbf{v}, \quad \mathbf{v}\sim U(-\Delta \mathbf{v},\Delta \mathbf{v}) \label{eq:dynamics}
\end{align}
where $\mathbf{x}\in\mathcal{X}\subset \mathbb{R}^d$ is the system state, the noisy observation is $\hat{\mathbf{x}}\in \mathbb{R}^d$, $\mathbf{u}\in \mathcal{U}\subset \mathbb{R}^d$ is the control input action and $\mathbf{w},\mathbf{v} \in\mathbb{R}^d$ are the process and measurement noises. $U$ denotes the uniform distribution, with a finite support defined by $\Delta \mathbf{w}$ and $\Delta \mathbf{v}$. Specifically, similar to {\cite{herbert2017fastrack}} we make the assumption that the underlying unknown dynamics of the system is uniformly continuous, bounded and Lipschitz continuous. Additionally, as mentioned earlier, we make an assumption that approximation error, uncertainty, and non-determinism in the system could be explained via noise with finite support \cite{Wenhao2019}. Thus, we can simplify the robot dynamics $\dot{\mathbf{x}}\in\mathbb{R}^d$ as stochastic control-affine single integrator dynamics of the form $\dot{\mathbf{x}} = \mathbf{u} + \mathbf{w}$, where $\mathbf{w}$ accounts for both the model non-linearity and any model uncertainties. This also allows us to make locally linear approximations of the dynamics over the control input. In the context of a quadrotor, the virtual control inputs provided through the single integrator dynamics are mapped to the corresponding non-linear physical model inside the simulation. 

\subsection{Safety Constraints}
Given our application domain of drone-racing, we propose safety constraints based on rich representations common in robotics. Our obstacle model is a static model and is represented through a distance transform inspired by the Euclidean signed distance field (ESDF). We define three regions for every obstacle $i$ with a pose denoted as $\mathbf{p}_i$, that are subsets of the 3D metric space: $\Omega_i^-$ inside the obstacle, $\Omega_i^+$ outside the obstacle and $\partial\Omega_i$ as its border. For any point in 3D space $\mathbf{x}$ that is the position of the robot, we define a custom distance function to obstacle $i$ as follows:
\begin{align}
d(\mathbf{x}, \mathbf{p}_i)&={\begin{cases}
\min_{\mathbf{y}\in\mathbf{\partial\Omega_i\cup\Omega_i^-}} \norm{\mathbf{x}-\mathbf{y}}, & \mathbf{x}\in(\Omega_i^+\cup\partial\Omega_i) \\
        -1, & \mathbf{x}\in\Omega_i^- 
\end{cases} } \label{eq:dist_func}
\end{align}

Under the assumption that a robot cannot physically be inside an obstacle, i.e. for every state $\mathbf{x}\notin\Omega_i^-$ and for every obstacle $i$, there are properties of the distance function defined in (\ref{eq:dist_func}) that are useful in defining the safety barrier. In particular, we make the following observation:
\\
\begin{remark}
$d(\mathbf{x},\mathbf{p}_i)$ where $\mathbf{x}\notin\Omega_i^-$ is Lipschitz continuous, differentiable almost everywhere and bounded under a finite support.\\
\label{remark:d_continuous}
\end{remark}

We define a state $\mathbf{x}\notin\Omega^-$ to be safe with respect to an obstacle $i$ with the pose $\mathbf{p}_i$ if the following conditions hold:
\begin{align}
h_{i}^s(\mathbf{x})&=d^2(\mathbf{x},\mathbf{p}_i)-\mathbf{R}^2,\qquad &\mathbf{x}\notin\Omega_i^- \label{eq:h^s_i}\\
\mathcal{H}_{i}^s&=\{\mathbf{x}\in \mathbb{R}^{d}:h^s_{i}(\mathbf{x})\geq 0\}, \qquad &\mathbf{x}\notin\Omega_i^-
\label{eq:H^s_i}
\end{align}

The set $\mathcal{H}_{i}^s$ indicates the set of states that are safe with respect to the obstacle $i$, where $\mathbf{R}$ is a buffer safety radius. Naturally, the condition of $\mathbf{x}\notin\Omega_i^-$ ensures valid robot states lie only outside obstacles. For our application in drone-racing, we consider square gates as obstacles (see Fig. \ref{fig:sdf_map}). Additionally, considering the finite boundary of (\ref{eq:dist_func}) and the fact that all our obstacles are the same, we precompute the signed distance field using (\ref{eq:dist_func}) for a set of sampled poses of the robot within a region of interest relative to the gate, thus creating a distance map. 

\begin{figure*}[htp]
   \begin{tabular}{p{0.7\textwidth}p{0.17\textwidth}}
    \hspace{0.2mm} 
    \includegraphics[scale=0.8]{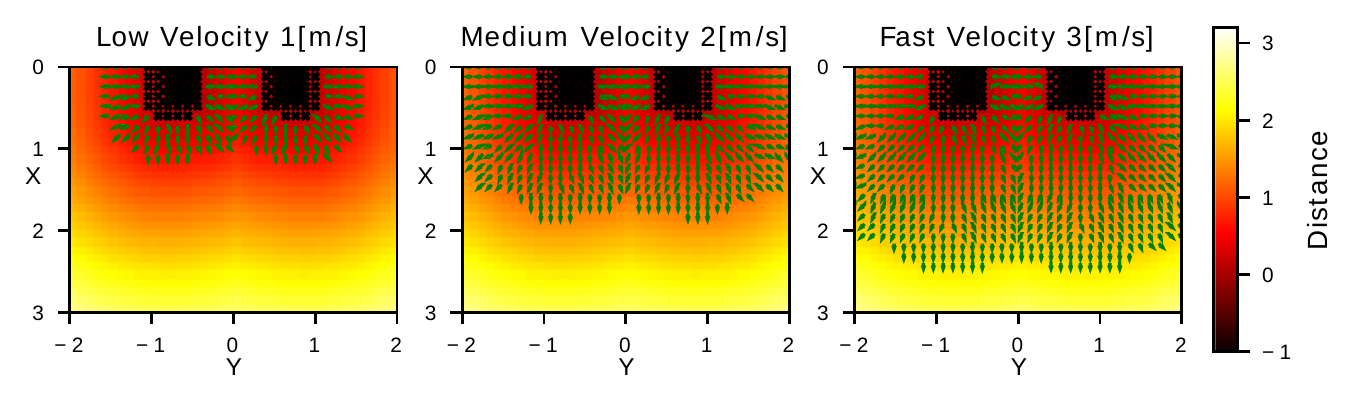}
    \label{fig:sdf_map_uncertainty11}&
    \multirow{2}{\hsize}[24mm]{
    \includegraphics[scale=0.27]{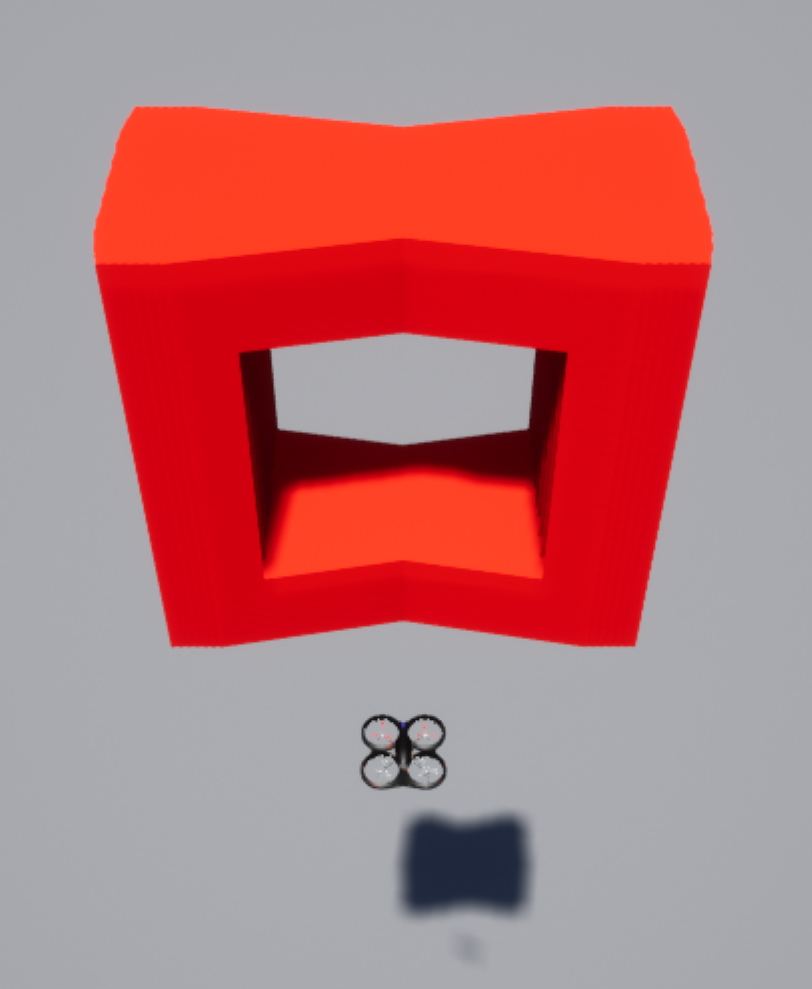}\label{fig:sdf_map_uncertainty12} }
    \\[-4mm]
    \includegraphics[scale=0.8]{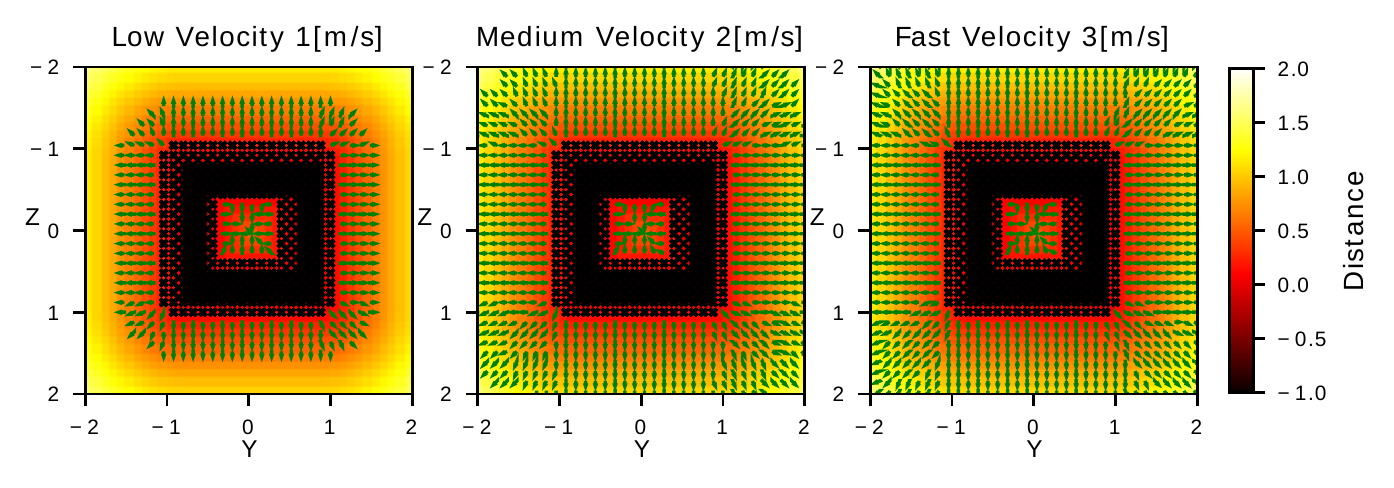}\label{fig:sdf_map_uncertainty13}&
    \end{tabular}
    \caption{The distance map defined by (\ref{eq:dist_func_star}), by $\mathbf{XY}$ and $\mathbf{YZ}$ planes (upper row and lower row). The reader is guided to zoom-in the figures for details: The green arrows direct the safest action as computed by the SBC and providing the safety constraints, and the black crosses indicate positions that are unsafe in every direction. The actions sampled for every position and for every angle are $|\mathbf{u}|=velocity$. The right image shows the new gate's structure, inflated by including all the possible positions the gate can take while considering uncertainty. Applying safety constraints affects the modifications to the original controller and consequently leads to more restrictions to it, as can be seen compared to Fig. \ref{fig:sdf_map}. }
    \label{fig:sdf_map_uncertainty}
\end{figure*}

\subsection{Safety Under Uncertainty}
In most of the deep control scenarios, the state variable used to define and evaluate safety is latent. Consequently, we assume that a state estimation routine is available that would provide the system state with a bounded error. In our work on drone-racing, we use a Variational Autoencoder (VAE) based module to estimate the pose of the gates. This estimation not precise but considered to have finite support. To address safety under the uncertainty arising due to such state estimation, we perform worst-case safety computation. Formally, we define a new distance function as follows:
\begin{align}
{d^*}(\mathbf{x}, \mathbf{p_i})&=\begin{cases}
\min_{\mathbf{p}^*_i\in\mathbf{P}_i} d({\bf x},\mathbf{p}^*_i), &\quad\{\mathbf{x}\in(\Omega_i^+\cup\partial\Omega_i), \\
&\quad \forall \mathbf{p}^*_i\in\mathbf{P}_i\}\\
-1, &\quad \mathbf{x}\in\Omega_i^-
\end{cases} \label{eq:dist_func_star}
\end{align}
where $\mathbf{P}_i$ is the set of points that is occupied when the obstacle $i$ is replicated at all possible positions within the error threshold of the predicted pose.\\
\begin{remark}
We can represent a new obstacle with pose $\hat{\mathbf{p}}_i$ that comprises of all possible positions $\mathbf{p^*}_i\in\mathbf{P}_i$ such that ${d^*}(\mathbf{x}, \mathbf{p}_i) = {d}(\mathbf{x}, \hat{\mathbf{p}}_i)$.\\
\label{remark:new obstacle}
\end{remark}
This new obstacle allows us to consider the worst-case scenario under gate pose estimation uncertainty and can be tackled using the same safety definition as in (\ref{eq:H^s_i}). A new corresponding distance map can also be precomputed. This method simplifies the way to provide safety under uncertainties as the basic underlying fabric stays unchanged. Fig. \ref{fig:sdf_map} and Fig. \ref{fig:sdf_map_uncertainty} show the distance maps with and without considering uncertainty, and the difference in the measurements of the obstacle. Now the safe set considering worst-case scenarios under uncertainty can be defined by:
\begin{align}
\mathcal{H^*}^s_i&=\{\mathbf{x}\in \mathbb{R}^{d}:{h^*}^s_i(\mathbf{x})\geq 0\},  \notag \\ 
\mbox{ where }
&{h^*}^s_i(\mathbf{x})={d^*}^2(\mathbf{x},\mathbf{p}_i)-\mathbf{R}^2,\mbox{ and } \mathbf{x}\notin\Omega_i^-
\end{align}

It is easy to show using Remark (\ref{remark:new obstacle}) that there exists an equivalent safety set that considers the original ESDF using the newly constructed obstacle with pose  $\hat{\bf p}_i$:
\begin{align}
\mathcal{H}^s_i&=\{\mathbf{x}\in \mathbb{R}^{d}:h^s_i(\mathbf{x})\geq 0\},  \notag\\
\mbox{ where }
&h^s_i(\mathbf{x})=d^2(\mathbf{x},\hat{\mathbf{p}}_i)-\mathbf{R}^2,\mbox{ and } \mathbf{x}\notin\Omega_i^-
\end{align}

\section{SAFETY BARRIER CERTIFICATES UNDER UNCERTAINTY}
Barrier certificates, or barrier functions, are used to ensure that robots remain in safe sets for all time. Controllers are expected to satisfy the barrier certificates while taking control actions that are as close as possible to the nominal action. For the discussion in this section, we assume the perspective of the robot that is running the perception-action loop, thus expressing the obstacle poses relative to the pose of the robot. This allows for a simplification of the notation for the distance function from $d(\mathbf{x},\mathbf{p})$ to $d(\mathbf{p})$. Under the assumption that at least one obstacle is in the field of view of the robot at any time, we simplify the notation and represent the safety set and constraints as a function of the next obstacle pose $\mathbf{p}\in\mathcal{P}\subset \mathbb{R}^d$ relative to the robot. The set $\mathcal{P}$ is again defined by all states that correspond to the center of the robot being outside the obstacle. The safety set in the new simpler representation is defined similarly to (\ref{eq:H^s_i}):
\begin{align}
\mathcal{H}^s &= \{\mathbf{p}\in \mathcal{P}\quad |\quad {h^s}(\mathbf{p})\geq 0\} \notag\\
\mbox{ where }
&{h^s}(\mathbf{p})=d^2(\mathbf{p})-\mathbf{R}^2,\mbox{ and } \mathbf{p}\in \mathcal{P} \label{eq:H_s_uncertainty}
\end{align}

Based on the theory of Zeroing Control Barrier Functions (ZCBF) and SBC, some conditions need to be applied to the controller $\mathbf{u}\in\mathcal{U}$ to guarantee forward invariance of the safety set. A continuously differentiable function $h^s: \mathcal{P}\rightarrow \mathbb{R}$ is a ZCBF, and the admissible control space can be defined as:
\begin{align}
S(\mathbf{p})&=\{\mathbf{u}\in\mathcal{U}\;|\;\dot{h}^s(\mathbf{p})+\kappa(h^s(\mathbf{p}))\geq 0\},\;\mathbf{p}\in \mathcal{P}
\end{align}
Any Lipschitz continuous action $\mathbf{u}\in S(\mathbf{p})$ guarantees that the set $\mathcal{H}^s$ is forward invariant. Considering the extended class-$\mathcal{K}$ function as $\kappa(h^s(\mathbf{p}))=\gamma h^s(\mathbf{p})$ for $\gamma>>0$, and based on the admissible control space, the SBC that defines the constraints can be formulated as:
\begin{align}
\mathcal{B}^s(\mathbf{p})&=\{\mathbf{u}\in \mathbb{R}^{d}:\dot{h}^s(\mathbf{p})+\gamma h^s(\mathbf{p})\geq 0\},\;\mathbf{p}\in \mathcal{P}
\label{eq:SBC}
\end{align}

\begin{remark}
For all the positions occupied by the robot where the distance map is differentiable, and assuming the initial position is collision-free, it can be shown that the constrained control space described by (\ref{eq:SBC}) induces a linear constraint over the robot controller. Proof and further discussion can be found in \cite{Wenhao2019}.\\
\end{remark}
We recall here that as we pre-compute a distance map for ${h^s}({\bf p})$ over a grid of sampled poses, it is possible to determine the relevant constraints efficiently at runtime. 

\begin{figure*}[htp]
    \begin{tabular}{cc}
        \includegraphics[width=0.46\textwidth]{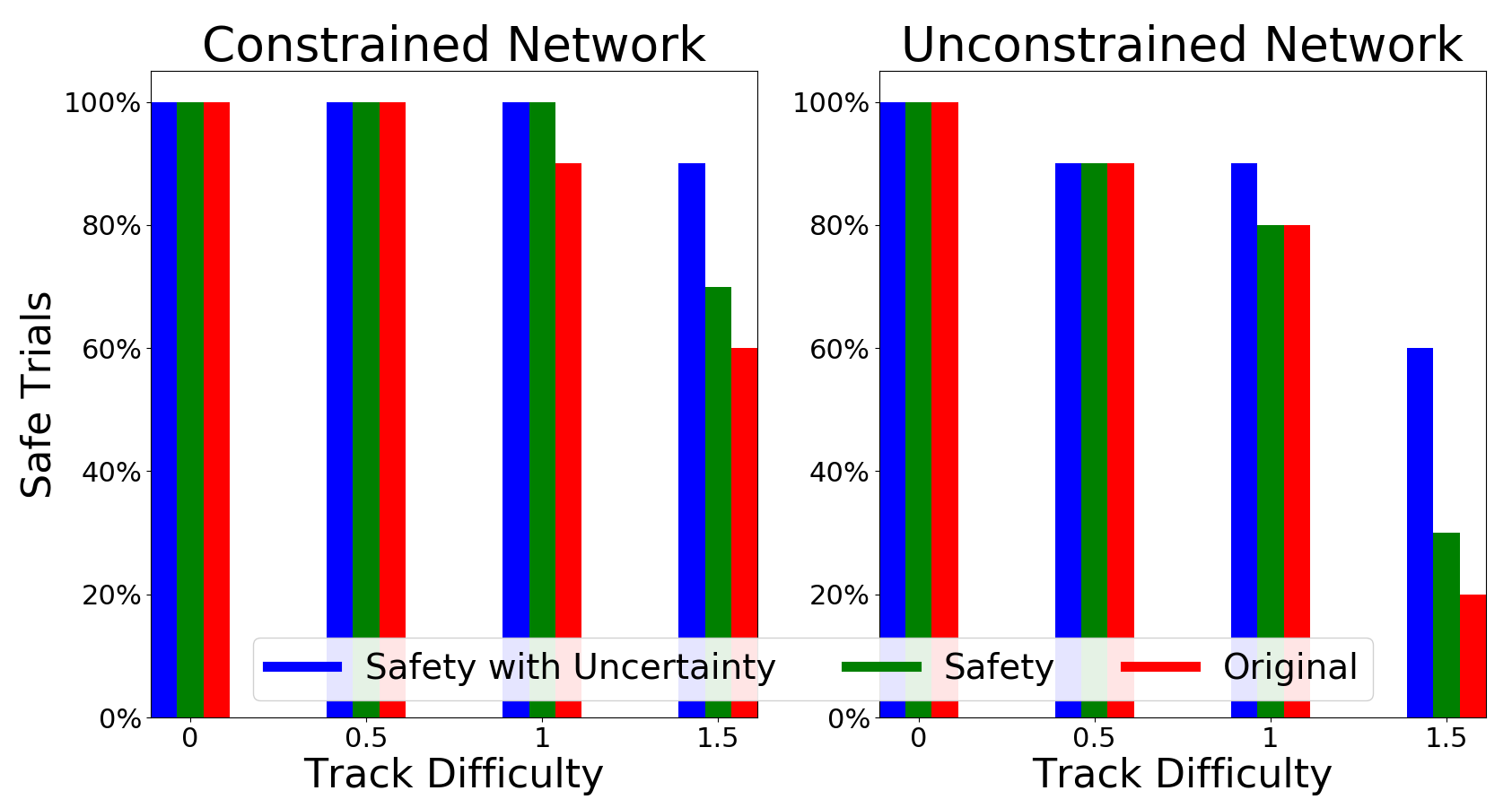} &
        \includegraphics[width=0.46\textwidth]{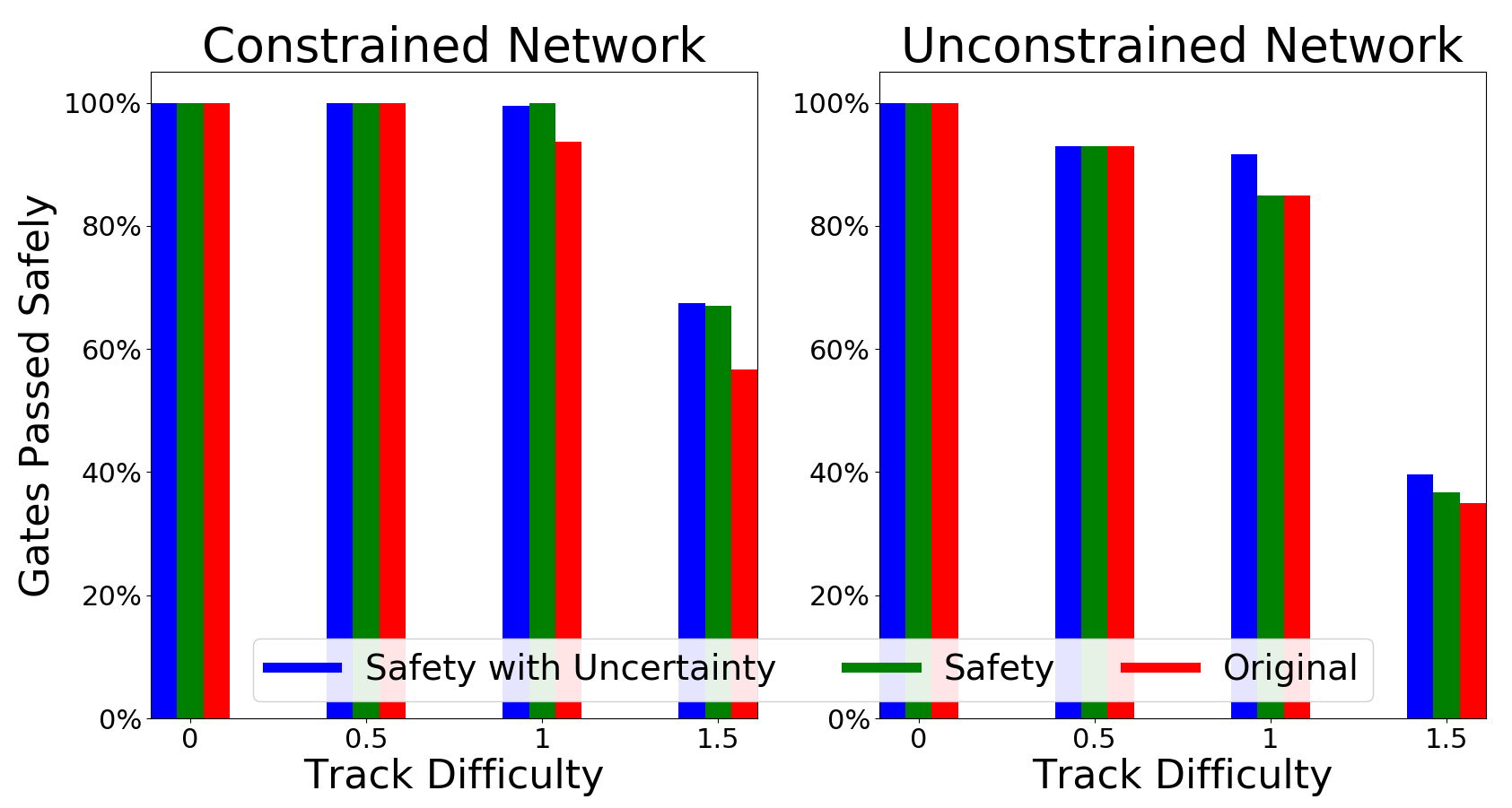} 
        \\\footnotesize{(a) Performance on Safety} & \footnotesize{(b) Performance on Task Success}\\
    \end{tabular}
    \caption{Performance comparison on (a) safety and (b) success metrics for different difficulty levels. We observe that the proposed framework provides better performance on safety, especially when considering the uncertainty in gate position, while still performing better than original baselines on the task. The effects are more pronounced when the difficulty level is higher.}
    \label{fig:experiments}
\end{figure*}

In order to ensure ${h^s}(\mathbf{p})$ defined in (\ref{eq:H_s_uncertainty}) is continuously differentiable, we can use a smooth approximation to (\ref{eq:dist_func}) (for example using the softmax trick). In our experiments, we simply work with ESDFs noting that the regions of non-differentiability (with respect to a drone gate's shape) arise only at the places where the vehicle is guaranteed to be safe by a wide margin. Second, we transform the VAE's estimated gate position coordinates from spherical to Euclidean coordinates, where the quadrotor's yaw angle is equal to the predicted angle with respect to the obstacle. These actions help prove the continuity of the derivative of $h^s({\bf p})$. Thus we can formally rewrite $\mathcal{B}^s$ as:
\begin{align}
\mathcal{B}^s(\mathbf{p})&=\Big\{\mathbf{u}\in \mathbb{R}^{d}:2d(\mathbf{p})\left(\frac{\partial{d(\mathbf{p})}} {\partial{\mathbf{x}}}\right)^T\mathbf{\dot{x}}\notag \\ 
 & +\gamma{(d^2(\mathbf{p})-\mathbf{R}^2)}\geq 0\Big\},\;\mathbf{p}\in \mathcal{P} \label{eq:B_s_p}
\end{align}

\begin{figure*}[htp]
    \begin{tabular}{p{0.4\textwidth}p{0.5\textwidth}}
        \includegraphics[scale=0.29]{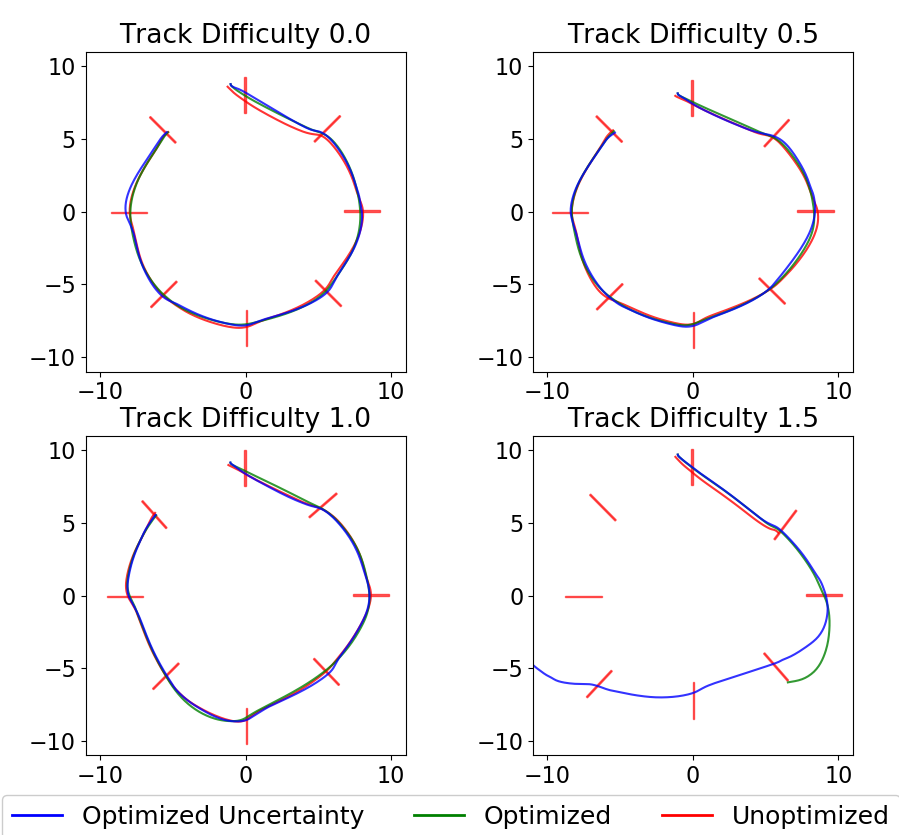}&
        \includegraphics[scale=0.3]{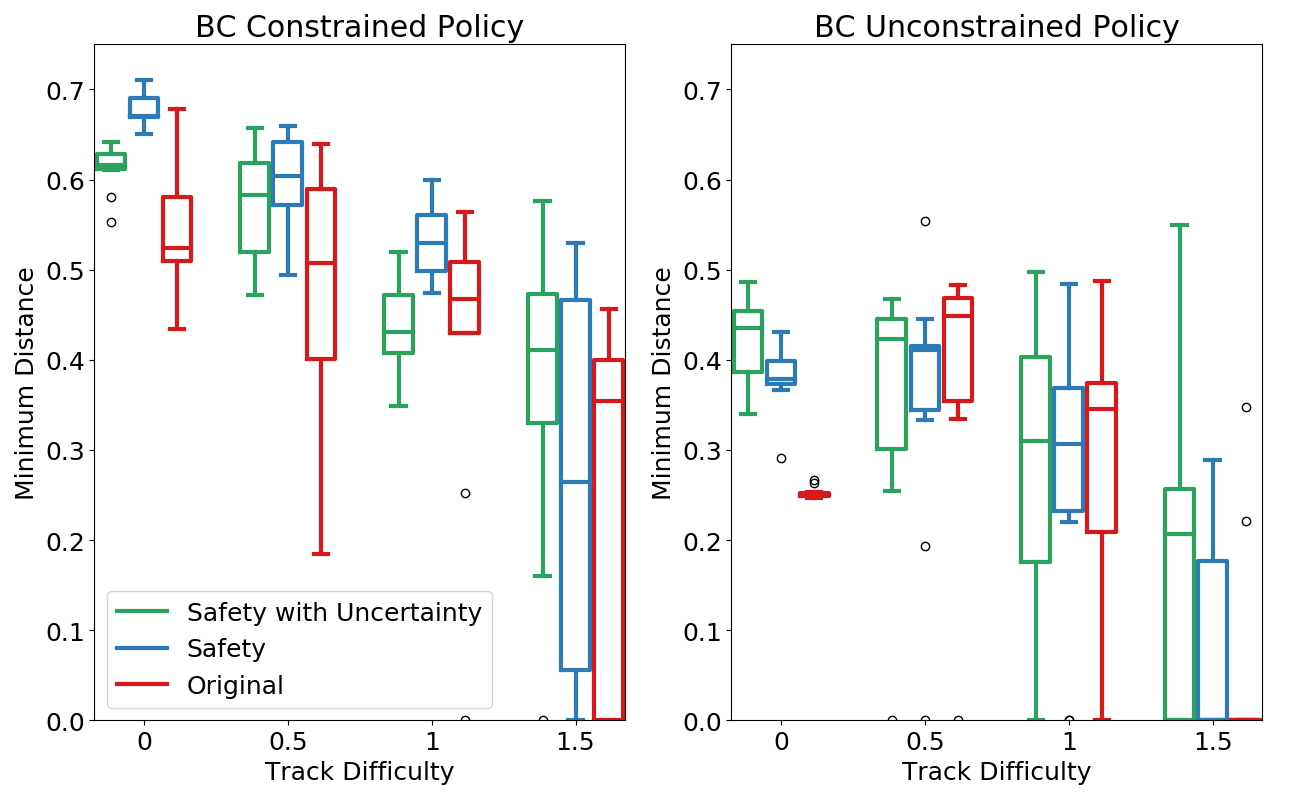} 
        \\
        \hspace{25pt}\footnotesize{(a) Trajectories on four track difficulty levels} &\hspace{45pt}\footnotesize{(b) Minimum distance - Averaged over track difficulty level}
    \end{tabular}

    \caption{(a) Sample trajectories from a trial for the original policy and the safety policies. (b) Boxplots summarizing the closest distance a quadrotor got to any gate, based on (\ref{eq:dist_func}), over the 10 experiments and for each difficulty level. The horizontal line indicates the median, while the boundaries of the boxes denote the 25\% and 75\% quantile levels, respectively. The whiskers correspond to the most extreme data points not considered outliers, and the outliers are plotted individually as `o'. For difficult tracks, the proposed framework leads to a better separation between the gates and the quadrotor.}
    \label{fig:risk}
\end{figure*}

\begin{figure*}[htp]
    \centering
    \includegraphics[scale=0.7]{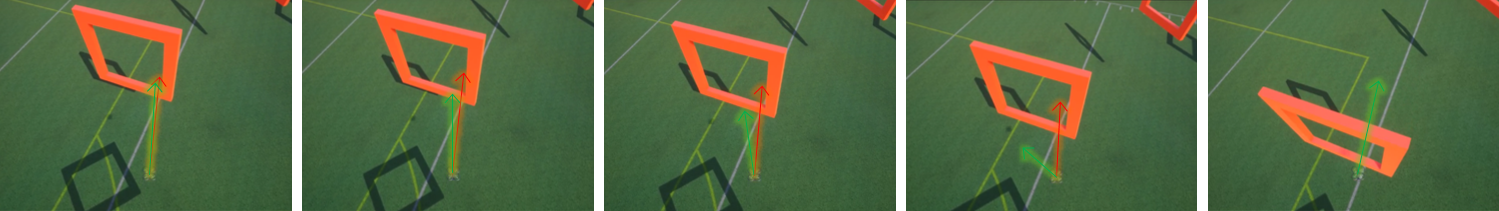}
    \caption{An example of our proposed framework in simulation, as the quadrotor was moving towards the gate. The figures are arranged from left to right with respect to the quadrotor's movement. In each figure, the red arrow indicates the original deep policy control, and the green arrow shows the safe action as inferred by the framework, while considering the uncertainty in the gate position.}
    \label{fig:paths_viz}
\end{figure*}

Similar to precomputing the distance map, it is also possible to precompute its gradient $\frac{\partial{d(\mathbf{p})}} {\partial{\mathbf{x}}}$. Further, under our assumed locally linear dynamics (\ref{eq:dynamics}), we can write $\mathbf{\dot{x}}$ in (\ref{eq:B_s_p}) to be $\mathbf{u} + \mathbf{w}$. Given that $\mathbf{w}$ has finite support ($|\mathbf{w}| \leq \Delta\mathbf{w}$) and $d(\mathbf{p})$ is Lipschitz continuous, we can compute a constant $C \geq 2d(\mathbf{p})(\frac{\partial{d(\mathbf{p})}} {\partial{\mathbf{x}}})^T\mathbf{w}$, thus resulting in a linear constraint on $\mathbf{u}$ that guarantees the inequality in  (\ref{eq:B_s_p}).
Finally, we formulate our safety problem as a Quadratic Program (QP) to minimally change the action if needed, i.e. modify the original control action if it is found to violate the safety constraints. Formally, we solve the following program using the safety constraints defined in (\ref{eq:H_s_uncertainty}) and the SBC (\ref{eq:B_s_p}):
\begin{align}
    \min_{\mathbf{u}^*\in \mathbb{R}^{d}} &\quad \norm{\mathbf{\tilde{u}}-\mathbf{u}^*}^2 \\
     \text{s.t.}
     &\quad \mathbf{p}\in \mathcal{H}^s \\
     &\quad \mathbf{u}^*\in\mathcal{B}^s(\mathbf{p})\\
    &\quad \norm{\mathbf{u}^*}\leq \alpha
\end{align} 
where $\alpha$ is the boundary of the controller action, $\mathbf{\tilde{u}}$ is the original deep policy control and the safe action is denoted by $\mathbf{u}^*$.
In practice, considering worst-case safety leads to a difference in the modifications to the original controller, and it is more restricted near the obstacle (for example see in Fig. \ref{fig:sdf_map} and Fig. \ref{fig:sdf_map_uncertainty}).

\section{EXPERIMENTS AND RESULTS}
We performed experiments to verify the robustness of the method and understand its limitations via a drone-racing simulation built on top of AirSim \cite{Shah_2017}. Each experiment comprised of a quadrotor navigating through a set of ten racing tracks for three laps. Each track was around 50m in length, built by eight gates positioned randomly. Each experiment was associated with one of four difficulty levels (ranging from 0 to 1.5 with a step size of 0.5), defined by the maximum offset between the centers of two consecutive gates, where a larger offset requires more maneuvering to stay on track.

We use two key metrics for evaluation: safety and the ability to solve the given tasks successfully. A trial consists of maneuvering through three consecutive laps of the track, and it is defined as {\em safe} when the quadrotor stays collision-free over the entire trial. The percent of gates negotiated safely through a trial is a measure of {\em success} on the task. We wish to explore if the proposed framework allows us to be safe while still being competitive, as defined by the success criterion. 

For the perception module and baseline control policies, we use the networks from \cite{bonatti2019learning}: a variational auto-encoder (VAE-constrained) that predicts next gate poses and Behavior Cloning (BC) policies constrained and unconstrained, which are the best performing networks for control in the mentioned work. We compare both deep control policies with our proposed safety framework with and without uncertainty (corresponding to gate pose localization). Our uncertainty estimation is based on the errors in gate pose estimation computed empirically by \cite{bonatti2019learning}.

Fig. \ref{fig:experiments} shows the performances of both baselines when augmented with our optimization method, by safety and success metrics. In our experiments, the success rate seems to be almost similar for all methods, with a slight advantage for the safety method considering uncertainty. As the track difficulty increases to 1.5, we observed the safety performance of the original policies deteriorate drastically, while that of the safety policies decreases slower. We observe that the best safety rates were achieved when considering uncertainty.

We evaluated the experiments also by a distance metric defined in (\ref{eq:dist_func}). For every trial, we recorded the minimum distance between the quadrotor and the next gate at each time step, which indicates how close the quadrotor was to a possible collision. If a trial ended in collision, then the score is zero. Fig. \ref{fig:risk}b shows the minimum distance values seen, averaged for each difficulty level. The results show that our proposed safety method considering uncertainties achieves the best performance overall. While we sometimes observe lesser minimum distances when considering uncertainty, this can be attributed to the fact that under uncertainty, the obstacle is artificially inflated to a larger size. A visualization of safe control commands and trajectories are shown in Fig. \ref{fig:risk}a and Fig. \ref{fig:paths_viz}, applied to the BC unconstrained network. In Fig. \ref{fig:risk}a, we show the differences between the original policy trajectories and the safety controls with and without considering uncertainties. The trajectories are almost the same for the first three difficulty levels, but when the difficulty level increases to 1.5, then the only safe trajectory is when considering safety, which leads out of track. For the same track, the original policy and the safety method lead to a collision with the second and fourth gates, respectively. A detailed control visualization is shown in Fig. \ref{fig:paths_viz}, where the actions of the original policy are violating the safety constraints and would lead to a possible collision with a gate, whereas the safety method with uncertainty computes a collision-free action.

We have observed a few limitations of the proposed method in our experiments. For example, when the angle between the current gate and next gate was too sharp but still in the field of view of the quadrotor, occasionally, all methods caused a collision with the current gate. Another issue we encountered was when the quadrotor starts a trial facing a gate's pole, and in close proximity. In this situation, the quadrotor most of the time collided with the gate for all the methods, which could have been because of significant noise in gate estimated position and estimation errors exceeding the worst-case values considered. One way to overcome such an issue is to consider optimization for the next two gates instead of only one.

\section{CONCLUSIONS AND FUTURE WORK}
We have presented a framework for safe deep control policies for the task of drone-racing. At the heart of our method are safety barrier certificates, used to minimally change the controller to ensure forward invariance of safety. The main idea to overcome uncertainty in obstacle position is considering the worst case in error threshold of predicted obstacle pose and building a pre-computed distance map through Euclidean signed distance field. Our experiments show that using our proposed method is elevating the safety rate of deep control policies, while still achieving competitive results. Future work includes investigating a prediction process of more than one gate position. We would also be interested in exploring the use of this method during training time of deep control policies, to balance safety and performance before execution.

\addtolength{\textheight}{0cm}   




\section*{ACKNOWLEDGMENT}

We would like to thank Ratnesh Madaan and Rogerio Bonatti for their inputs regarding the baseline perception and control policies; as well as Matthew Brown and Nicholas Gyde for their help with the simulations.


\bibliographystyle{IEEEtran}
\bibliography{IEEEabrv, root}

\end{document}